\documentclass[10pt,twocolumn,letterpaper]{article}

\usepackage{cvpr}              % To produce the CAMERA-READY version
\def\cvprPaperID{52} % *** Enter the CVPR Paper ID here
\def\confName{CVPR}
\def\confYear{2022}
\usepackage[pagebackref]{hyperref}       % hyperlinks
\usepackage{url}            % simple URL typesetting
\usepackage{booktabs}       % professional-quality tables
\usepackage{amsfonts}       % blackboard math symbols
\usepackage{nicefrac}       % compact symbols for 1/2, etc.
\usepackage{microtype}      % microtypography

\usepackage{graphicx}
\usepackage{comment}
\usepackage{amsmath,amssymb} % define this before the line numbering.
\usepackage{xcolor}
\usepackage[inline]{enumitem}
\usepackage{epstopdf}
\usepackage{caption}
\usepackage{subcaption}
\usepackage[font={small}]{caption}
\usepackage{tabularx}
\usepackage{booktabs}
\usepackage{multirow}
\usepackage{pifont}
\usepackage{arydshln}
\usepackage{microtype}
\usepackage{wrapfig}

\newenvironment{tight_itemize}{
\begin{itemize}[leftmargin=10pt]
  \setlength{\topsep}{0pt}
  \setlength{\itemsep}{2pt}
  \setlength{\parskip}{0pt}
  \setlength{\parsep}{0pt}
}{\end{itemize}}

\usepackage{array}
\newcolumntype{L}[1]{>{\raggedright\let\newline\\\arraybackslash\hspace{0pt}}m{#1}}
\newcolumntype{C}[1]{>{\centering\let\newline\\\arraybackslash\hspace{0pt}}m{#1}}
\newcolumntype{R}[1]{>{\raggedleft\let\newline\\\arraybackslash\hspace{0pt}}m{#1}}

\renewcommand{\eqref}[1]{Eq.~(\ref{#1})}

\newcommand{\tabref}[1]{Table~\ref{#1}}
\newcommand{\figref}[1]{Figure~\ref{#1}}

\newcommand{\D}{\mathbb{D}}

\newcommand{\Y}{\mathbb{Y}}

\newcommand{\E}{\mathcal{E}}
\newcommand{\FTN}{\mathcal{F}}
\newcommand{\Dec}{\mathcal{G}}
\newcommand{\Prob}{\mathcal{P}}
\newcommand{\Disc}{\mathcal{D}}
\newcommand{\Loss}{\mathcal{L}}

\usepackage{amsmath}

\DeclareMathOperator*{\argmin}{arg\,min}

\usepackage{IEEEtrantools}

\begin{document}

\title{Cluster-to-adapt: Few Shot Domain Adaptation \\ for Semantic Segmentation across Disjoint Labels} % Replace with your title
\author{Tarun Kalluri \quad Manmohan Chandraker \\
University of California San Diego
% {\tt\small sskallur@eng.ucsd.edu}
}

\maketitle

\begin{abstract}
Domain adaptation for semantic segmentation across datasets consisting of the same categories has seen several recent successes. However, a more general scenario is when the source and target datasets correspond to non-overlapping label spaces. For example, categories in segmentation datasets change vastly depending on the type of environment or application, yet share many valuable semantic relations. Existing approaches based on feature alignment or discrepancy minimization do not take such category shift into account. 
In this work, we present Cluster-to-Adapt (C2A), a computationally efficient clustering-based approach for domain adaptation across segmentation datasets with completely different, but possibly related categories. 
We show that such a clustering objective enforced in a transformed feature space serves to automatically select categories across source and target domains that can be aligned for improving the target performance, while preventing negative transfer for unrelated categories. 
We demonstrate the effectiveness of our approach through experiments on the challenging problem of outdoor to indoor adaptation for semantic segmentation in few-shot as well as zero-shot settings, with consistent improvements in performance over existing approaches and baselines in all cases. 
\end{abstract}

\section{Introduction} 

In this work, we address the problem of knowledge transfer across domains with disjoint labels for semantic segmentation.
In spite of massive strides in computer vision performance using deep learning~\cite{lecun2015deep}, models trained on a large-scale labeled dataset are not guaranteed to generalize to data that lies outside the training distribution.
This difficulty is amplified for applications like semantic segmentation, where collecting pixel level labeled data for all geographies, environments and weather conditions is restrictive, expensive or simply not feasible due to many practical and social implications~\cite{kalluri2019universal}. 

\begin{figure}
    \centering
    \includegraphics[width=0.5\textwidth]{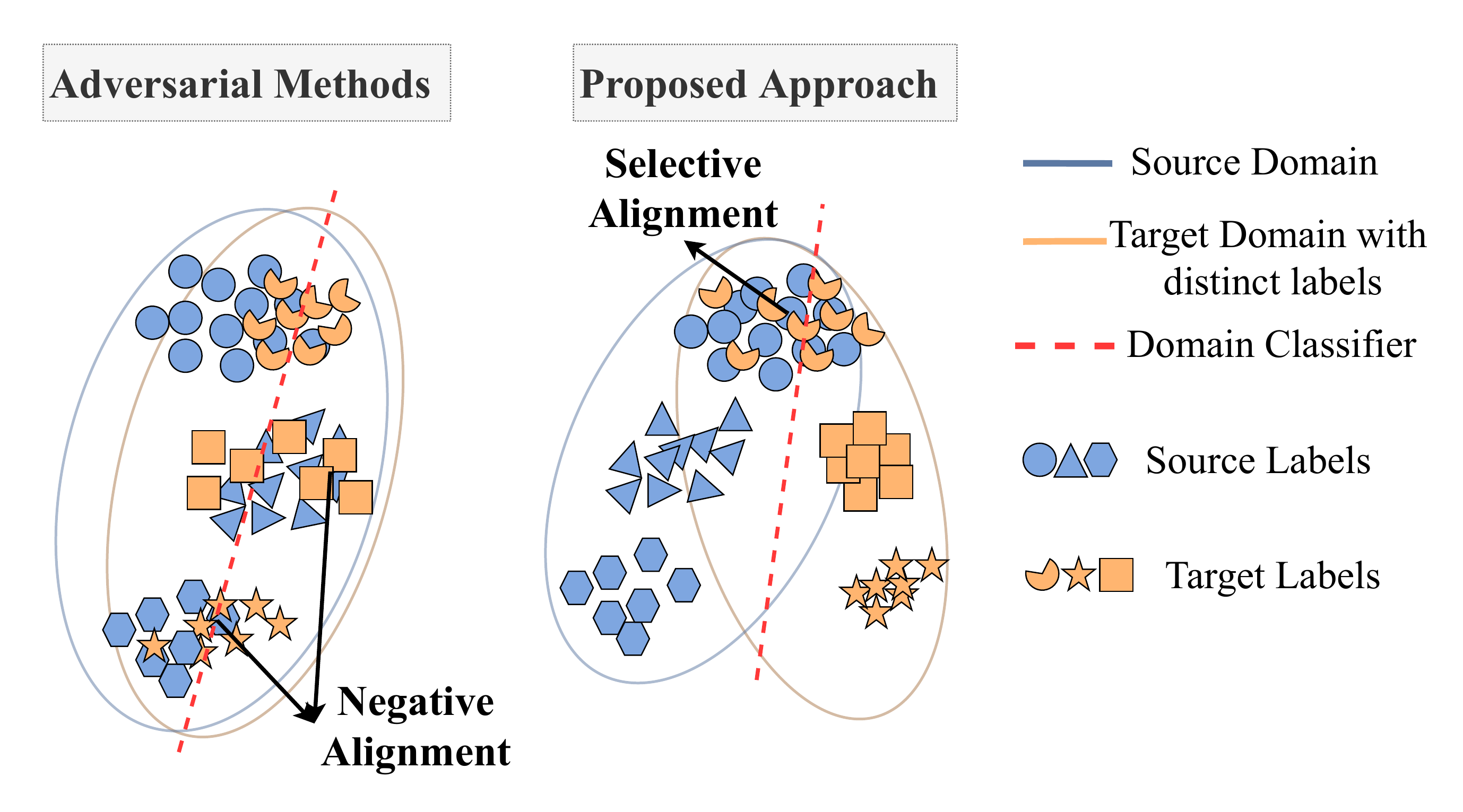}
    \captionsetup{width=0.5\textwidth}
    \caption{We illustrate the major challenge in our setting arising due to negative transfer. In our setting where source and target do not contain shared label space, traditional adversarial methods relying on global adaptation might lead to unrelated classes aligning with each other. In contrast, our method encourages only related classes to align with each other (selective alignment), while preventing negative transfer among unrelated classes.}
    \label{fig:illustration}
\end{figure}

Unsupervised domain adaptation emerged as a feasible alternative to transfer knowledge from a labeled source domain to unlabeled target domains by minimizing some notion of divergence between the domains~\cite{long2017deep,long2015learning,sun2016deep,ganin2016domain,tzeng2017adversarial,bousmalis2016domain}.
Prior works in domain adaptation are based on a global distribution alignment objective, assuming that the source and target datasets share the same label space so that domain alignment would invariably result in learning transferable feature representations. 

% \begin{figure}[t]
    
%         \centering
%         \includegraphics[width=.85\textwidth]{eccv2020kit/scripts/figures/dataset_relation.pdf} % second figure
%         % itself
%         \caption{\textbf{Overview of proposed approach} To transfer knowledge from synthetic outdoor images to real indoor images, we use unlabeled images from real outdoor scenes as an intermediate domain, and propose a combination of adaptation and clustering objectives to perform knowledge transfer.}
%         \label{fig:dataset_relation}
    
% \end{figure}

% \begin{wrapfigure}{r}{0.5\textwidth}
%     %
%         \centering
%         \includegraphics[width=.48\textwidth]{eccv2020kit/scripts/figures/dataset_distances-NeuRIPS_al_fig.pdf} % second figure
%         % itself
%         \caption{\textbf{Illustration of Selective Alignment using C2A:} While previous adversarial approaches give rise to negative transfer effects, using C2A, we perform selective alignment across source and target datasets, which helps in aligning only related categories while maintaining sufficient distance between unrelated ones.}
%         \label{fig:illustration}
%     %
% \end{wrapfigure}
% %%%%%%%%%%%%%%%%%~~~~~~~~~~~~~~~~~~~~~~%%%%%%%%%%%%%%%%%
% %%

\begin{figure*}[!t]
    \begin{center}
    \includegraphics[width=0.9\textwidth]{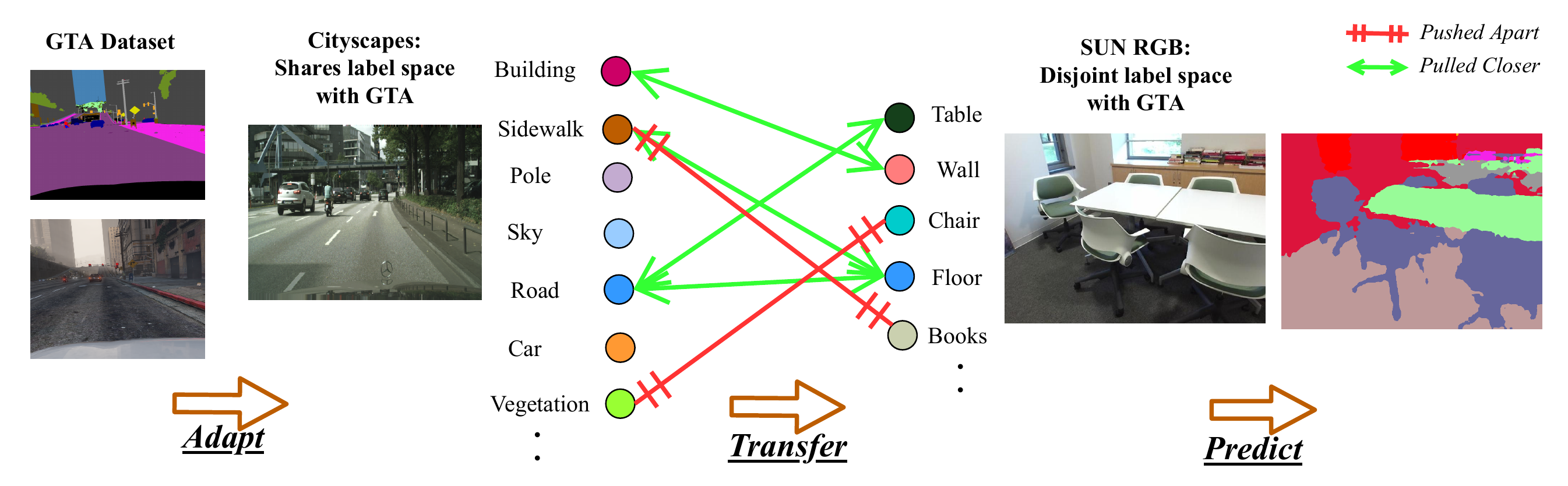}
    \end{center}
    \captionsetup{width=\textwidth}
    \caption{Overview of the proposed adaptation across disjoint labels using unlabeled bridge domains. Note that the source and target domains have completely disjoint label spaces here, which is a very realistic setting for real world transfer learning scenarios. We tackle this challenging setting in our problem using an unlabeled \textit{bridge} domain. In this case, where synthetic driving data is acting as the labeled source domain with real indoor scenes as the target domain, real driving scenes can be used as the bridge as it shares properties with both of these, enabling few shot adaptation to the target domain. }
    \label{fig:dataset_relation}
  \hfill
%   \begin{minipage}{0.35\linewidth}{
%     \begin{center}
%     \includegraphics[width=\textwidth]{eccv2020kit/scripts/figures/dataset_distances-NeuRIPS_al_fig.pdf}
%     \end{center}
%     \captionsetup{width=\textwidth}
%     \subcaption{}
%     \label{fig:illustration}
%   }
%   \end{minipage}
%   \caption{\protect\subref{fig:dataset_relation}To transfer knowledge from synthetic outdoor images to real indoor images, we use unlabeled images from real outdoor scenes as an intermediate domain, and propose a combination of adaptation and clustering objectives to perform knowledge transfer. \ref{fig:illustration} While previous adversarial approaches give rise to negative transfer effects, using C2A, we perform selective alignment across source and target datasets, which helps in aligning only related categories while maintaining sufficient distance between unrelated ones.}
    
    % \vspace{-1em}
\end{figure*}

In many cases, the source and target labels might be completely distinct and share only high level geometric and semantic relationships. This makes it hard, yet necessary in few-shot settings, to perform useful knowledge transfer. In particular we show this in case of adaptation between outdoor datasets, where synthetic datasets are readily available, and indoor scenes, where we have few labeled data and it is considered difficult to render or maintain synthetic datasets. 
To address this challenging setting of outdoor to indoor adaptation, we propose a novel framework for adaptation across disjoint labels. For disjoint labels, we posit that a more suitable objective is to achieve \textit{domain invariance} with respect to related categories and \textit{domain equivariance} with respect to unrelated categories between source and target, thus avoiding negative transfer.
For example, the categories that frequently occur in an indoor environment like wall, floor, ceiling and chair are completely distinct from any outdoor categories, yet we show how we can leverage useful discriminative information through implicit geometric and semantic correspondences.

In practice, the distribution shift between such source and target domains arise from both low-level (lighting, contrast, object density etc.) and high-level (category, geometric orientation, pose etc.) variations~\cite{bousmalis2016domain}. To ease this extreme case of adaptation, we introduce an additional unlabeled auxiliary domain, which shares properties with both the source and target datasets and would act as a bridge to improve the adaptation. 
For instance, adaptation from synthetic outdoor to real indoor datasets can benefit from unlabeled images from real outdoor scenes, as explained in~\figref{fig:dataset_relation}.

To automatically discover related and unrelated categories across datasets, we propose a novel clustering based alignment approach called Cluster-to-adapt (C2A). C2A stems from the intuition that related categories from source and target should lie close to each other in the feature space for effective knowledge transfer. We realize this during training through a deep constrained clustering framework by posing the alignment as a clustering objective in a transformed feature space, which would force related categories to group close to each other while leaving room for unrelated categories to form independent clusters, as shown in~\figref{fig:illustration}.

In summary, we make the following contributions.

\begin{tight_itemize}
    \item A novel cluster-to-adapt (C2A) approach is proposed to effectively perform category level adaptation between semantic segmentation datasets with disjoint labels, using an intermediate domain with shared properties. As such, we address the most general domain adaptation setting of knowledge transfer across datasets with completely distinct label spaces for semantic segmentation.
    \item We make use of a computationally efficient clustering framework that helps in reducing the distance between related categories across datasets during training while preventing negative alignment between unrelated categories.
    \item We demonstrate through empirical results that our proposed C2A approach consistently outperforms related approaches and baselines in fewshot as well as zeroshot settings for adaptation between outdoor and indoor segmentation datasets.
\end{tight_itemize}

\section{Related Work}

\paragraph{\textbf{Unsupervised Domain Adaptation (UDA)}}

UDA is used to transfer knowledge from a large labeled source domain to an unlabeled target domain. Large body of works that perform adaptation from labeled source to unlabeled target rely on adversarial generative~\cite{bousmalis2017unsupervised,sankaranarayanan2018generate,murez2018image,hoffman2017cycada, zou2019confidence} or  discriminative~\cite{ganin2016domain,tzeng2017adversarial,tzeng2015simultaneous,hoffman2016fcns} approaches to learn domain agnostic feature representations. 
A common assumption in most of these approaches is that the source and target label spaces completely overlap, so that a classifier learnt on the source domain can be directly applied on the target data. However, in most real world applications this assumption is invalid, and in most general case, the categories might be completely different. Very few works exist which address this more challenging setting.
Previous works like open set adaptation~\cite{saito2018open}, partial set adaptation~\cite{cao2018partial,cao2019learning} and universal adaptation~\cite{you2019universal, kalluri2019universal} assume some degree of label overlap, \cite{luo2017label} performs adaptation between distinct label spaces with few target labeled data using pairwise similarity constraints, while \cite{sohn2018unsupervised} addresses adaptation for verification tasks which is different from our focus on semantic segmentation. Similarly, more recent works for domain adaptation suited for semantic segmentation tasks \cite{lv2020cross, musto2020semantically, yang2020fda, mei2020instance, wang2020classes, araslanov2021self} achieve state of the art results for the case of completely overlapping label spaces in the source and target domains, and are not applicable in our setting of outdoor to indoor adaptation.
In contrast to these existing works, we propose an efficient method to align only visually similar features across source and target domains which can have completely non-intersecting label spaces without re-annotation~\cite{lambert2020mseg}, while preventing potential negative transfer, specifically suited to cross domain semantic segmentation.

\paragraph{\textbf{Deep Clustering}} Although clustering algorithms like k-means~\cite{macqueen1967some} are extremely useful in automatically discovering structure from unlabeled data~\cite{aggarwal2014data}, they work directly on the high dimensional input space like images which is often ineffective for classification. Recent works propose jointly learning a suitable feature representation of data along with clustering assignments. For example, \cite{hsu2017learning} uses pairwise similarity based constraints, while~\cite{han2019learning} uses self-training objective on the cluster assignment scores to successfully perform unsupervised transfer across categories from the same domain. Other works make use of deep clustering to learn more discriminative clusters~\cite{xie2016unsupervised} useful for classification, or as suitable pretext tasks in self supervised learning~\cite{caron2018deep,caron2019unsupervised,yan2019clusterfit}. While deep clustering based approaches have been previously applied in the case of unsupervised category discovery~\cite{han2019learning}, we extend this idea to additionally account for the domain shift between the source and target datasets.

\textcolor{black}{Also, note that many prior works that use clustering for adaptation consider the classical setting of completely matching source and target domains \cite{DBLP:conf/wacv/ToldoMZ21, DBLP:conf/eccv/WuHLUGLD18, DBLP:journals/corr/abs-2012-04280} or partial overlap in open world setting \cite{DBLP:conf/cvpr/GongCPLCLDG21}, and hence use clustering as a means to achieve one-to-one alignment between source and target. In contrast, we use clustering to selectively align source and target across completely disjoint label spaces. }

\section{Framework}

We now explain our proposed approach, which addresses the most general case of knowledge transfer between domains with different, and non-overlapping label spaces. 
Denote using $\D_s$ the completely labeled source domain data with label space $\Y_s$, where $\D^s \sim p_s$(source distribution). The labeled target domain data is denoted by $\D_t$, with label space $\Y_t (\neq \Y_s)$, and $\D_t \sim p_t$(target distribution).
% , so they are essentially different tasks following the nomenclature in~\cite{pan2009survey}. 
We assume that a small subset $\D_t^l$ of the target data is labeled, for learning some task specific information like classifier boundaries, and the rest $\D_t^u$ as unlabeled, making our setting that of \emph{few-shot adaptation across domains with disjoint labels}. We denote this small fraction of labeled samples by $\sigma = |\D_t^l|/(|\D_t^u| + |\D_t^l|)$. Following the nomenclature of \cite{pan2009survey}, we henceforth call this as cross-task adaptation and the source and target as different tasks. In section \ref{sec:experiments}, we show results varying $\sigma$ from $0.01$ to $0.3$.
% We are mainly interested in cases when $ \sigma \ll 1 $.
In our case, the domain gap between the target data and source data comes from two factors, namely domain shift due to $p_s \neq p_t$ as well as label shift due to $\Y_s \neq \Y_t$. Furthermore, we do not assume any partial overlap between the label spaces unlike other partial or open set adaptation approaches which makes our setting more challenging. To ease the adaptation process across these widely different datasets, we introduce another completely unlabeled auxiliary domain $\D_a$, which serves as a useful bridge between the source and target datasets. For example, $\D_a$ could share \textit{task/content} properties with $\D_s$ and \textit{style} properties with $\D_t$. We show how to best exploit this completely unlabeled intermediate domain to achieve our primary goal of learning transferable features from source to target.

\begin{figure*}[!t]
  \centering
  \includegraphics[width=\linewidth]{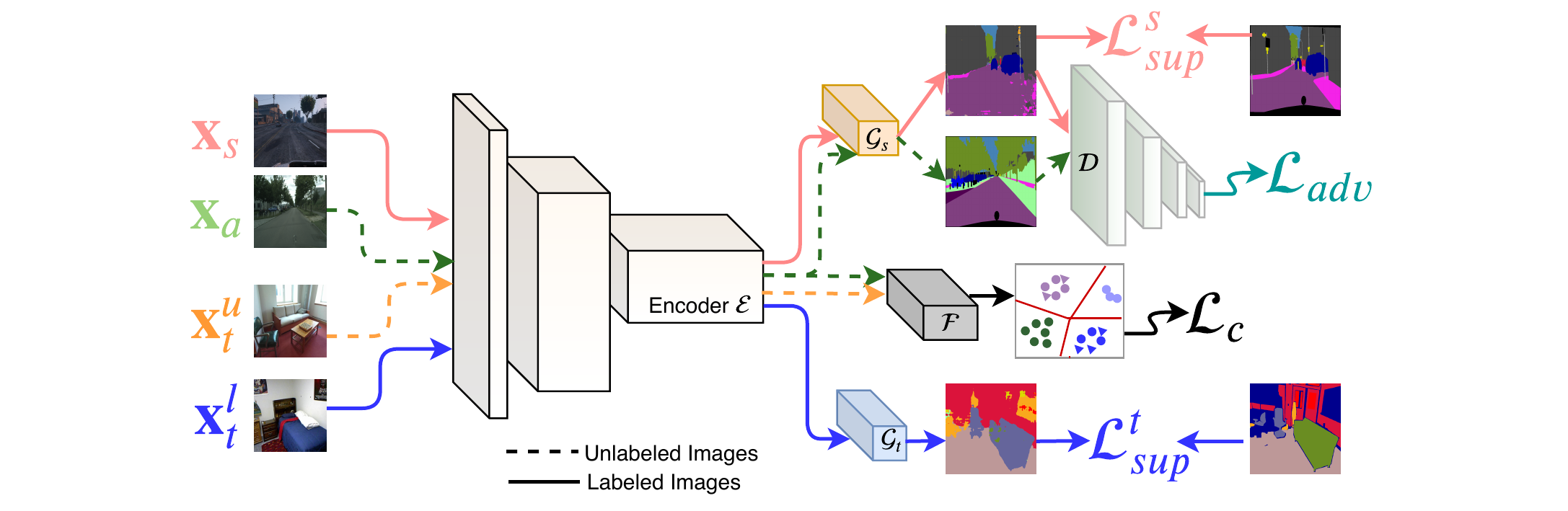}
    \caption{\textbf{C2A: Proposed Architecture.} The labeled data from the source ($\D_s$, in red) and target ($\D_t^l$, in blue) are used to train the task specific decoders using supervised losses $\Loss_{sup}^s$ and $\Loss_{sup}^t$. The unsupervised images from auxiliary domain ($\D_a$, in green) is used in adversarial adaptation from source to intermediate domain using loss $\Loss_{adv}$. The target unlabeled data ($\D_t^u$, in orange), along with $\D_a$, is used to compute the clustering loss $\Loss_c$ in a feature space transformed by $\FTN$. The encoder $\mathcal{E}$ is common for all the images.
  }
  \label{fig:main_schematic}
\end{figure*}

% \begin{figure}
%   \begin{minipage}[c]{0.67\textwidth}
%     \includegraphics[width=\linewidth]{eccv2020kit/scripts/figures/architecture.pdf}
%   \end{minipage}\hfill
%   \begin{minipage}[c]{0.3\textwidth}
%     \caption{\textbf{C2A: Proposed Architecture.} The labeled data from the source ($\D_s$, in red) and target ($\D_t^l$, in blue) are used to train the task specific decoders using supervised losses $\Loss_{sup}^s$ and $\Loss_{sup}^t$. The unsupervised images from auxiliary domain ($\D_a$, in green) is used in adversarial adaptation from source to intermediate domain using loss $\Loss_{adv}$. The target unlabeled data ($\D_t^u$, in orange), along with $\D_a$, is used to compute the clustering loss $\Loss_c$ in a feature space transformed by $\FTN$. The encoder $\mathcal{E}$ is common for all the images.
%   } \label{fig:main_schematic}
%   \end{minipage}
% \end{figure}

\noindent {\bf Overview} We present the overview of the proposed network architecture in~\figref{fig:main_schematic}. We use a shared encoder $\E:(H,W,3) \rightarrow (H',W',f_d)$ across all the datasets which aggregates spatial features across multiple resolutions of the input image $x$, and outputs a downsampled encoder map $\E(x)$. $f_d$ is the size of features in the encoder map. Since shallow level features are known to be more task agnostic and transferable~\cite{yosinski2014transferable}, a shared encoder helps us to learn generic features useful across source and target datasets. Task specific decoders $\Dec_s$ and $\Dec_t$ then upsample the output of the encoder and compute class assignment probabilities for each pixel of the input image over the label space $\Y_s$ and $\Y_t$ respectively. Individual decoders for source and target helps us to make predictions over respective label spaces. The supervised loss computed using the labeled data from source and target datasets is given by  
\begin{equation}
    \Loss_{sup} = \Loss_{sup}(\D_s) + \Loss_{sup}(\D_t^l),
    \label{eq:supervised_loss}
\end{equation}
where
\begin{IEEEeqnarray}{rCl}
    \Loss_{sup}(\D_s) &=& \frac{1}{N_s} \nonumber \\ &&\sum_{\substack{(x,y) \in \{\D_s\}}} \frac{-1}{H W} \sum_{h,w} \log(\Dec_s(\E(x))^y(h,w))) \IEEEeqnarraynumspace
\end{IEEEeqnarray}
and $N_s$ is the number of labeled samples in the source dataset, and $H,W$ are the height and width of the output feature map respectively. The target supervised loss $\Loss_{sup}(\D_t^l)$ is defined similarly.  
% We then calculate the supervised loss $\Loss_{sup}$ using these predictions and the ground truths available from source data.  

% However, training $\E$ and $\Dec$ on labeled source data alone would not perform well on target data due to \textit{within task domain gap}. 
% To address the within-task domain gap challenge, 
%
Next, we decouple the source to target alignment into two different objectives. The first is a \textit{within task} alignment between $\D_s$ and $\D_a$, and the second is the \textit{cross task alignment} objective between $\D_a$ and $\D_t$, as explained next. 
\subsection{Within Task Domain Alignment}

We introduce the within task alignment objective between the source and intermediate domains $\D_s$ and $\D_a$. We assume that the domains share the same label space and exhibit only low-level differences, and use an adversarial alignment strategy using a domain discriminator $\Disc$. 

Following the idea presented in~\cite{tsai2018learning}, we send the output probability maps $\Prob_s(x) = \Dec_s(\E(x))$ to the discriminator $\Disc$ as opposed to the encoder maps. This helps in better within-task alignment for pixel level prediction tasks and, as we found out, faster convergence during training. We train the discriminator $\Disc:(H,W,|\Y_s|) \rightarrow \{0,1\}$ which takes as input the output map of the generator, to output the probability of the map coming from source data. The generator is then trained to produce outputs from $\D_a$ which are good enough to trick the discriminator into classifying them as coming from source. This alternative min-max optimization would then result in domain invariant output maps leading to successful feature alignment. 
The adversarial loss, using LS-GAN~\cite{mao2017least}, is given by
\begin{equation}
    \Loss_{adv} = \mathrm{E}_{x \sim \D_a} (\Disc(\Prob_s(x)))^2 
\end{equation}
and the discriminator objective $\Loss_D$ is given by 
\begin{equation}
    \Loss_{\Disc} = \mathrm{E}_{x \sim \D_s} (\Disc(\Prob_s(x)))^2 + \mathrm{E}_{x \sim \D_a} (\Disc(\Prob_s(x))-1)^2
\end{equation}

Although we use this adversarial adaptation strategy for within-task alignment, we note that our method can also be applied in combination with any other adaptation strategy based on generative modeling or distribution matching~\cite{murez2018image,yang2020fda,vu2019advent} for within task alignment.

\subsection{Cross Task Semantic Transfer}
% \vspace{-.5em}

Training $\E$ and $\Dec$ with task specific supervised loss and adversarial alignment losses alone is insufficient to transfer useful semantic content to target dataset, since we do not explicitly transfer any semantic relations between the tasks. Naive adversarial training of yet another discriminator to distinguish outputs from two tasks would not work well, as we only want categories that share semantic cues to align with each other (selective alignment) as opposed to global alignment (\figref{fig:illustration}). Luo et. al.~\cite{luo2017label} propose using an entropy minimization objective after computing pairwise similarity of the features, but computing such pairwise similarity is computationally infeasible for pixel level prediction tasks.
Towards this goal, we propose a novel deep clustering based approach, which lies at the core of our approach.

\paragraph{\emph{\textbf{Constrained Clustering Objective}}}
% Distance metric. 
Following the assumption that deep features form discriminative clusters in the feature space useful for classification tasks~\cite{chapelle2005semi}, we believe that better knowledge transfer would happen across tasks if the features of categories which share semantic information also form coherent clusters closer to each other. A major challenge with incorporating this constraint in deep neural networks is the lack of information regarding the correspondence between categories of the datasets useful in preventing negative transfer effects. 
We use a clustering based objective to discover the similarity across categories, and enforce the clustering constraint by performing k-means clustering of the feature vectors. This encourages the features corresponding to similar categories across tasks to form a single ``meta-cluster", while leaving room for unrelated categories to form independent clusters.  

We first pass the outputs of the shared encoder $\E$ through a feature transfer module $\FTN:(H',W',f_d) \rightarrow (H',W',f_e)$, where $f_e$ is the feature dimension in the transformed space. $\FTN$ is necessary because the features learnt specific to a task might not be suitable for cross-task semantic transfer directly in the feature space. A learnable transformation function would, instead, find the best subspace amenable for alignment. Also, since $f_e \ll f_d$, the feature transformation would result in efficient computation of centers and similarity metrics for k-means. We formulate our constrained clustering objective using the cross-entropy loss, given by 
\begin{equation}
    \Loss_c = \sum_{x \in  \{\D_a , \D_t^u\}} \sum_{v_j \in \FTN(\E(x))} - \log (\max_k p(\mu_k | v_j))
    \label{eq:clustering_loss}
\end{equation}

\noindent where $p(\mu_k|v_j)$ is the probability score that a feature vector $v_j$ belongs to a cluster with center $\mu_k$, and
\begin{equation}
    p(\mu_k | v_j) \propto \exp \left( \frac{v_j \cdot \mu_k}{||v_j||_2 ||\mu_k||_2} \right)
\end{equation}

\paragraph{\emph{\textbf{Avoiding Trivial Solution}}} Direct optimization of~\eqref{eq:clustering_loss} would quickly lead to a trivial solution where all the vectors are mapped to a single cluster.
We found that initializing the cluster centers using features computed from pretrained network on labeled target data alone ($ \D_t^l$, trained offline) would reduce this problem to a large extent. 
Additionally, we follow the idea proposed in~\cite{xie2016unsupervised}, and add a self-training constraint which encourages uniformity among the clusters and the cluster assignment probabilities are forced to be equal to an auxiliary target distribution. Specifically, we would like to have the target distribution $q(\mu_k | v_j)$ to hold the property that 
\begin{equation}
    q(\mu_k | v_j) \propto p(\mu_k | v_j) \cdot p(v_j | \mu_k)  \nonumber
\end{equation}

\noindent The first term on the RHS would improve the association of correct points to clusters, while the second term would discourage very large clusters. Applying bayes rule would give us the form of the target distribution as
\begin{equation}
    q(\mu_k | v_j) = \frac {{ p(\mu_k | v_j)^2}/ {\sum_j p(\mu_k | v_j)} } { \sum_{k'} { p( \mu_{k'} | v_j)^2} /{\sum_j p( \mu_{k'} | v_j)}   }.
\end{equation}

\noindent The constraint is now enforced in the form of a KL-Loss between the source distribution and the auxiliary target distribution. 
\begin{IEEEeqnarray}{rClCl}
    \Loss_{kl} &=& KL(p||q) \nonumber \\ &=& \sum_j \sum_k q(\mu_k | v_j) \log \left(\frac{q(\mu_k | v_j)}{p(\mu_k | v_j)} \right)
    \IEEEeqnarraynumspace
\end{IEEEeqnarray}

% \vspace{-1em}
\noindent The final training objective for the model can be summarised as follows,
\begin{eqnarray}
    \argmin_{\E,\Dec_s,\Dec_t,\FTN} & \,\,\, \Loss_{sup} + \lambda_{adv} \Loss_{adv} + \lambda_{c} (\Loss_{c} + \Loss_{kl}) \nonumber \\
    \argmin_{\Disc} & \Loss_{\Disc} 
\end{eqnarray}

\vspace{-1em}
\noindent where $\lambda_{adv}$ and $\lambda_c$ are the coefficients which control the relative importance of the adversarial loss and clustering loss respectively. The optimization is done by alternating between the two objectives within every iteration. A crucial factor in our method is the initialization of cluster centers before training, and we discuss our strategy followed for it next.

\subsection{Cluster Initialization}

The cluster centers are initialized using networks pretrained on the limited labeled data. Specifically, we use the same architecture as described in the paper to train a model on the labeled source data $\D_s$ as well as sparsely labeled target data $\D_t^l$ using pixel level cross entropy loss. We then pass the unlabeled images from the target $\D_t^u$ and collect all the encoder maps corresponding to all the images. Each encoder map is of size $(H/8 , W/8 , 2048)$ for our ResNet-101 backbone. To match the dimension of the FTN output, which is $128$ in our case, we apply PCA over these feature vectors to reduce their dimension. Then, a clustering is performed using the classical k-means objective with $K$ cluster centers, and the resulting centers are used to initialize $\mu_k's$ in the downstream adaptation approach. 

\paragraph{ \emph{\textbf{Efficient Computation of Centers}}} In traditional k-means, the centers $\mu_k$ are calculated using an iterative algorithm consisting of cluster assignment and centroid computation repeated until convergence. We mention a couple of issues persistent with this approach. Firstly, for dense prediction tasks like semantic segmentation, the encoder map consists multiple feature vectors which correspond to different patches of the input image. For example, an encoder map of size $(H',W')$ has $H' W'$ vectors of size $f_e$. Performing k-means over these vectors collected over all images over all the tasks would demand huge storage and computation requirements. Secondly, switching between gradient based training of network parameters and iterative computation of cluster centers after every few iterations would lead to an inefficient procedure that is not end-to-end trainable. To counter these limitations, we follow the idea proposed in ~\cite{han2019learning} and include $\mu_k$ as trainable parameters in the network, and update them after each iteration based on the gradients received from $\Loss_C$. 

\section{Experiments}
\label{sec:experiments}

% \begin{table}
%   \small
%   \centering
%   \resizebox{0.5\textwidth}{!}{
%   \begin{tabular}{@{} l c c c c c c c @{}} 
%     \toprule  \\[-1em]
%      \multirow{2}{*}{Method} &
%      \multicolumn{3}{c}{N=50}  && \multicolumn{3}{c}{N=100} \\ %&& \multicolumn{2}{c}{N=300} \\
%       \cmidrule{2-4} \cmidrule{6-8} %\cmidrule{8-9}
%     & CS &  CVD & Avg. && CS & CVD & Avg. \\%&& CS & CVD  \\
%     \midrule \\[-1.5ex]
%     Train on CS & 33.33 & 32.92 & 33.13 && 40.97 & 36.52 & 38.75 \\ 
%     Train on CVD & 19.47 & 42.81 & 31.14 && 22.20 & 50.02 & 36.11 \\ 
%     \midrule
%     Univ-basic & 32.82 & 48.56 & 40.69 && 36.04 & 51.90 & 43.97 \\ 
%     Univ-cross & 33.86 & 52.57 & 43.22 && 37.82 & 49.31 & 43.57 \\
%     Univ-full & 34.01 & 53.23 & \textbf{43.62} && 41.03 & 54.62 & \textbf{47.83} \\ 
%     % Univ-full ($K=3$) & \textbf{39.3} & \textbf{51.31} && \textbf{43.77} & \textbf{54.64} \\ 
%     \bottomrule \\[-1.5ex]
%   \end{tabular}}
%  \vspace{-0.5em}
%   \caption{\label{tab:few_shot_cs_cvd} mIoU values for universal segmentation using Cityscapes (CS) and CamVid (CVD) datasets with a Resnet-18 backbone. $N$ is the number of supervised examples available from each dataset. Bold entries have the highest \textit{average mIoU} across the datasets.}
%   \vspace{-1.2em}
% \end{table}

\begin{table*}[!t]
    %\vspace{-2em}
    \small
    \begin{center}
    \resizebox{0.7\textwidth}{!}{
    \begin{tabular}{l p{1.4cm}  p{1.4cm}  p{1.4cm} p{1.6cm}}
    & \multicolumn{3}{c}{$\xleftarrow{\text{Few-shot settings (our goal)}}$} & \\% & $\xrightarrow{\text{Full Supervision}}$ \\
    Method &                                        $\sigma$ = 0.01 &           $\sigma$ = 0.04 &           $\sigma$ = 0.1 &      $\sigma$ = 0.3 \\
           &                                        $N=50$&           $N=200$&           $N=500$&      $N=1500$ \\
    \midrule
    Target Labeled Only &                                   22.62 &             30.43 &             36.62 &         43.17 \\
    Fine Tune &                                     21.44 &             29.46 &             34.84 &         \textbf{44.10} \\ 
    \midrule
    AdaptSegNet*~\cite{tsai2018learning} &           25.20 &             32.51 &             36.9 &          43.83 \\
    LET*~\cite{luo2017label} &                       25.19  &             32.44  &             35.87&          42.96 \\
    \midrule
    \textcolor{black}{
    UnivSeg~\cite{kalluri2019universal}} & 22.21  & 31.32 & 36.08 & 42.10 \\
\textcolor{black}{AdvSemiSeg~\cite{hung2018adversarial} }& 24.72 & 33.22 & 38.46 & 45.10 \\
    \midrule
    % C2A (Ours, $\mathbf{\lambda_{adv}=0}$) &             24.10 &             32.22 &             35.89 &         43.08 \\
    C2A (Ours, $\mathbf{\lambda_c=0}$) &             24.10 &             32.22 &             35.89 &         43.08 \\
    C2A (Ours, full, K=10) &                               \textbf{25.98}$\pm 0.03$ &    \textbf{33.37}$\pm 0.07$ &    \textbf{37.41}$\pm 0.04$ & 43.16$\pm 0.03$ \\
    \end{tabular}
    }
    \end{center}
    \caption{\textbf{Few Shot Adaptation:} mIoU values on SUN-RGB validation set for the proposed few shot segmentation approach. $\sigma$ is the fraction of labeled examples from the total number of images from SUNRGB dataset. Note that our method particularly shows improvement in cases when the amount of labeled examples is very low. K is the number of clusters. (* denotes our extension of the existing works to suit our task.)
    \label{tab:main_results}
% *\textit{Modified to suit our setting}.
    }
\end{table*}

% We now validate the effectiveness of the proposed approach using empirical results on a challenging setting of adapting outdoor synthetic datasets like GTA~\cite{richter2016playing} and Synthia~\cite{ros2016synthia} to indoor real datasets like SunRGB~\cite{song2015sun}, using Cityscapes~\cite{cordts2016cityscapes} as the unlabeled intermediate domain. 
% We present more details regarding the choice of datasets, the network architecture, training details and metrics in the supplementary section. 
% We present details regarding the choice of datasets, the network architecture, training details and evaluation metrics, followed by our experimental results. 
\subsection{Datasets} 

For the source dataset $\D_s$, we use synthetic images from the driving dataset GTA~\cite{richter2016playing}. GTA consists of 24966 images synthetically generated from a video game consisting of outdoor scenes with rich variety of variations in lighting and traffic scenes. We also show results using the SYNTHIA-RAND-CITYSCPAES split from Synthia~\cite{ros2016synthia} dataset, which consists of 9600 synthetic images with labels compatible with Cityscapes. 
For the target dataset $\D_t$, we use real images from SUN-RGBD~\cite{song2015sun} consisting of images from indoor scenes. SUN-RGBD consists of 5285 training images and 5050 validation images containing pixel level labels of objects which frequently occur in an indoor setting like chair, table, floor, windows etc. We use the 13 class version from~\cite{mccormac2017scenenet}. The background class is ignored during training and evaluation.
% We believe that this choice of datasets is significant because knowledge transfer between indoor and outdoor images for semantic segmentation was never explored in literature before. 
Additionally, we use the 2975 training images from Cityscapes~\cite{cordts2016cityscapes} dataset, which consists of outdoor traffic scenes captured from various cities in Europe, as the unlabeled auxiliary domain $\D_a$. Cityscapes shares its semantic categories with GTA, so that the variation between $\D_s$ and $\D_a$ is only due to synthetic and real appearance, while $\D_s$ and $\D_t$ have many low-level as well as high-level differences. 
% The semantic categories are shared between Cityscapes and GTA.

% Furthermore, we also test out approach in a complementary, yet related, setting of zero shot adaptation. Specifically, we assume that we do not have any labeled or unlabeled images from the target dataset. Instead, we assume we have access to synthetically generated images that share categories with the target dataset. To validate this approach, we use all the labeled images from SceneNet~\cite{mccormac2017scenenet} as the target dataset, and show the generalization capabilities to datasets not seen before.
% We report all our results on the validation set of SUN-RGBD datasets. 

\subsection{Training Details}
We use the DeepLab~\cite{chen2017deeplab} architecture with a resnet-101 backbone for the encoder framework $\E$. For the task-specific decoder $\Dec$, we use an ASPP convolution layer followed by an upsampling layer. The architecture of discriminator $\Disc$ is similar to DC-GAN~\cite{radford2015unsupervised} with four $4 \times 4$ convolution layers, each with stride 2 followed by a leaky ReLU non-linearity. The feature transformation module $\FTN$ is a $1 \times 1$ convolution layer with output channels equal to the embedding dimension, which is fixed as $f_e=128$ for all the experiments. We use a default value for $\lambda_{adv}=0.001$. Following~\cite{xie2018learning}, to suppress the noisy alignment during the initial iterations, we set $\lambda_c = \tfrac{2}{1+e^{-10*\delta}}-1$, where $\delta$ changes from 0 to 1 over the course of training. The backbone architecture is trained using SGD objective, with an initial learning rate of $2.5 \times 10^{-4}$. For training the cluster centers, we follow a similar learning rate decay schedule, but start with a smaller learning rate of $2.5 \times 10^{-5}$. This is because the cluster centers are already initialized using networks trained on the labeled data, and we would ideally like the centers to not drift too far away from their initial values.

% We use $\lambda_{adv} = \lambda_c = 1$.

\vspace{-1em}
% \subsection{}
\paragraph{\emph{\textbf{Baselines and Ablations}}}
We provide ablation studies of the clustering module proposed in our approach and compare with the existing baselines. Specifically, we provide comparisons against the following.
\begin{enumerate*}[label=(\roman*)]
  \item \textbf{Target Labeled Only}: We train the segmentation encoder and decoder using only the limited labeled data from the target domain, $\D_t^l$, 
  \item \textbf{Finetune}: We use a model trained on source dataset $\D_s$ till convergence, and finetune it on the labeled target data,
  \item \textbf{Ours (C2A), $\lambda_c=0$}: Our cluster to adapt approach, without the clustering objective, and
  \item \textbf{Ours (C2A)}: our proposed approach with all the losses included.
\end{enumerate*}
\begin{table*}[!t]
\begin{center}
\small
\resizebox{0.96\textwidth}{!}{
\begin{tabular}{ l  l*{12}{c}  c c c c @{}}
    \toprule  \\[-1em]
    \rotatebox{0}{\footnotesize Method} & 
    \rotatebox{45}{\footnotesize bed} &
    \rotatebox{45}{\footnotesize books} & \rotatebox{45}{\footnotesize ceil.} & \rotatebox{45}{\footnotesize chair} & \rotatebox{45}{\footnotesize floor} & \rotatebox{45}{\footnotesize furn.} & \rotatebox{45}{\footnotesize objs.} & \rotatebox{45}{\footnotesize paint.} & \rotatebox{45}{\footnotesize sofa} & \rotatebox{45}{\footnotesize table} & \rotatebox{45}{\footnotesize tv} &
    \rotatebox{45}{\footnotesize wall} &
    \rotatebox{45}{\footnotesize win.} &&&
    \rotatebox{45}{\footnotesize pAcc.} &
    \rotatebox{45}{\footnotesize mIoU}
    \\ 
    \midrule \\[-1.5ex]
    &\multicolumn{17}{c}{\textbf{GTA to SunRGB}} \\
    \midrule \\[-1.5ex]
    Target Labeled & 36.47 &9.25 &27.15 &45.0 &71.13 &25.21 &6.76 &22.88 &25.86 &36.13 &0.0 &58.55 &31.16 &&& 65.02 & 30.43 \\
    Fine tune & 29.46 &8.09 &33.71 &44.33 &70.89 &23.75 &9.06 &\textbf{25.4} &24.3 &35.76 &0.0 &61.05 &25.5 &&& 64.60 & 29.46 \\ 
    % AdaptSegNet & 37.52 &15.04 &31.4 &47.2 &73.59 &26.23 &11.95 &22.34 &31.93 &36.88 &0.02 &62.13 &29.26 &&& 67.26 & 32.73 \\
    C2A [Ours] & \textbf{40.75} & \textbf{12.8} & \textbf{37.77} & \textbf{46.73} & \textbf{75.7} & \textbf{25.56} & \textbf{9.52} & 23.3 & \textbf{30.33} & \textbf{36.5} &0.0 & \textbf{61.51} & \textbf{33.42} &&&
     \textbf{66.71} & \textbf{33.37} \\
    \midrule \\[-1.5ex]
    &\multicolumn{17}{c}{\textbf{Synthia to SUNRGB}} \\
    \midrule \\
    Target Labeled & 36.47 &9.25 &27.15 &45.0 &71.13 &25.21 &6.76 &\textbf{22.88} &25.86 &36.13 &0.0 &58.55 &31.16 &&& 65.02 & 30.43 \\
    Fine tune & 31.87 & 9.76 & 34.98 & 43.89 & 71.44 & 25.67 &7.76 &15.74 &24.44 &36.12 &0.0 &61.30 &\textbf{32.98} &&& 65.08 & 30.46 \\ 
    % AdaptSegNet & 37.52 &15.04 &31.4 &47.2 &73.59 &26.23 &11.95 &22.34 &31.93 &36.88 &0.02 &62.13 &29.26 &&& 67.26 & 32.73 \\
    C2A [Ours] & \textbf{40.84} & \textbf{14.32} & \textbf{36.39} & \textbf{47.76} & \textbf{73.33} & \textbf{25.95} & \textbf{12.11} & 19.26 & \textbf{31.16} & \textbf{39.98} &0.0 & \textbf{63.92} & 32.77 &&&
     \textbf{67.37} & \textbf{33.68} \\
    \bottomrule
\end{tabular}}
\end{center}
\captionsetup{width=0.95\textwidth , font=footnotesize}
\caption{Classwise IoU values for the 13 classes in SUN RGB validation set compared against baselines for $\sigma=4\%(N=200)$. Classes like floor, wall and ceiling which share rich geometric and semantic properties with categories in outdoor GTA benefit the most, thus validating our approach of selective knowledge transfer across indoor and outdoor datasets. }  
\label{tab:class_wise_iou}
\end{table*}
%%%%%%%%%%%%%%%%%%%%%%%%%%%%

\vspace{-1em}
\paragraph{\emph{\textbf{Comparison with prior works}}} We reiterate the paucity of existing works which tackle the same setting as ours, making direct comparison hard. Many traditional adaptation methods prevalent in literature for segmentation~\cite{luo2019taking, hoffman2016fcns, sun2019not, vu2019advent} are not directly applicable in cases with disparate source and target label sets. Therefore, we compare against two competitive approaches that perform domain adaptation by extending them as follows.
% To provide comprehensive insights into the usefulness of our approach, we compare our work by extending a competitive approach that performs pixel level domain adaptation. 
\begin{enumerate*}[label=(\roman*)]
  \item \textbf{AdaptSegNet*}~\cite{tsai2018learning}: We choose \cite{tsai2018learning} as the backbone pixel level adaptation method for global adaptation across source and target datasets as it achieves high performance with a simple method. Since \cite{tsai2018learning} is not directly applicable to our case due to different labels spaces, we extend their method to perform feature space adaptation.
  \item \textbf{LET*}~\cite{luo2017label}: We compare against adaptation proposed in~\cite{luo2017label} using entropy minimization criterion. We extend it to suit the segmentation task by using class prototypes in the feature space instead of pairwise enumeration which keeps the computation feasible.
%   \item \textbf{Ours (C2A) , $\lambda_{adv} = 0$}: Our setting without using the intermediate domain $\D_a$.
\end{enumerate*}

% \begin{wrapfigure}{r}{0.5\textwidth}
% %   \captionsetup{width=\textwidth}

% \end{wrapfigure}

\vspace{-1em}
\paragraph{\emph{\textbf{GTA to SunRGB}}} 
% In ~\tabref{tab:main_results}, we provide ablations with varying the amount of supervision from the 5285 images in the SUN-RGB training set. 
We show in \tabref{tab:main_results} our results by varying the amount of supervision by choosing $|\D_t^l| = \{50,200,500,1500\}$ images which corresponds to $\sigma = \{1\% , 4\% , 10\% , 30\%\}$ respectively. Our method based on a novel clustering objective consistently outperforms other approaches by considerable margins, more so in cases when there is extreme scarcity of labeled data. We see upto $~15\%$ and $~10\%$ relative increase in mIoU for $\sigma=1\%$ and $\sigma=4\%$ respectively compared to training only on the labeled target dataset. It is also evident that the clustering loss $\Loss_c$ is important for the objective to successfully carry selective alignments from source and intermediate domains to the target domain, as seen from improvements in our results compared to prior works like \cite{tsai2018learning} and \cite{luo2017label}. We also observed that 30\% is already sufficient data for supervised fine-tuning to do well without any adaptation. 
% Similar observations have been made for other semi-supervised adaptation works~\cite{zhang2018fully}. 
In this work, our goal is focused on boosting the adaptation performance when enough labeled examples are not present in the target domain ($\sigma \lll 1$). 

% Proposed approach also exhibits remarkably significant gains compared to state of the art segmentation models based on discriminative adversarial objective like AdaptSegNet~\cite{tsai2018learning}. This is because direct adversarial approach would perform global distribution alignment, which is not always suitable in settings where the categories from the datasets do not overlap. In contrast, our constrained clustering based approach would positively transfer discriminative information across only a set of selected distinct yet related categories.
\begin{figure*}
% \vspace{-0.2em}
\centering
        \begin{subfigure}[b]{0.248\textwidth}
                \centering
                \includegraphics[width=\linewidth]{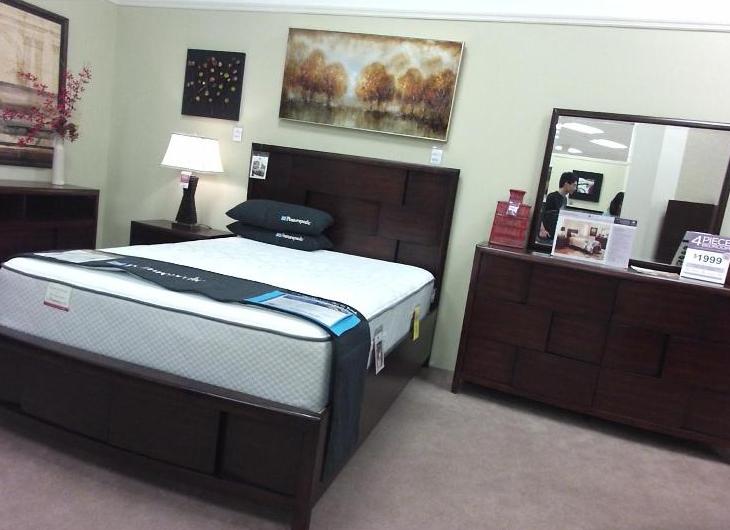}
                \vspace{-4em}
                % \caption{\label{fig:anue_image}}        
        \end{subfigure}\hfill
        \begin{subfigure}[b]{0.248\textwidth}
                \centering
                \includegraphics[width=\linewidth]{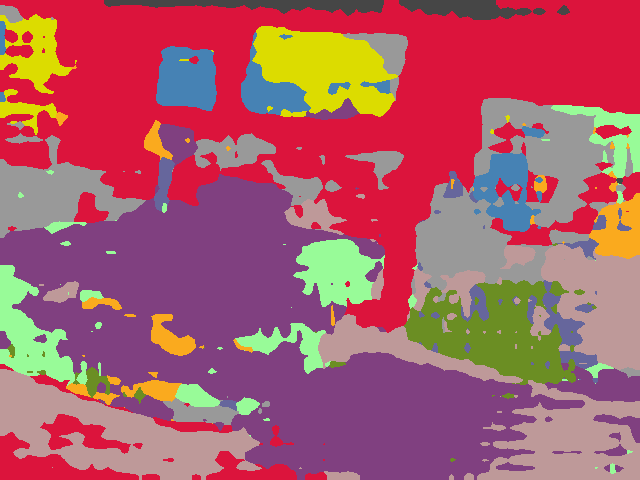}
                \vspace{-4em}
                % \caption{\label{fig:anue_bad}}
        \end{subfigure}\hfill
        \begin{subfigure}[b]{0.248\textwidth}
                \centering
                \includegraphics[width=\linewidth]{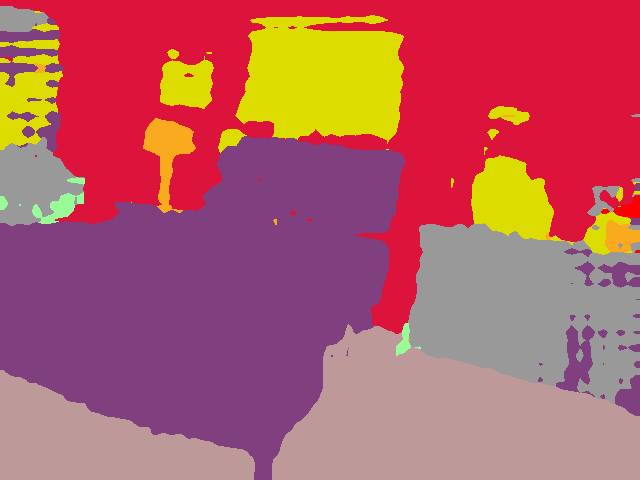}
                \vspace{-4em}
                % \caption{\label{fig:anue_good}}
        \end{subfigure}\hfill
        \begin{subfigure}[b]{0.248\textwidth}
                \centering
                \includegraphics[width=\linewidth]{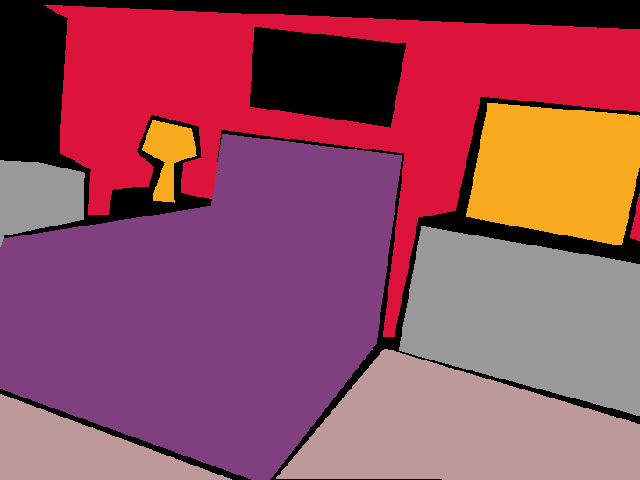}
                \vspace{-4em}
                % \caption{\label{fig:anue_label}}         
                
        \end{subfigure}\hfill
        ~
        \vspace{1em}
        \par\bigskip
                \begin{subfigure}[b]{0.248\textwidth}
                \centering
                \includegraphics[width=\linewidth]{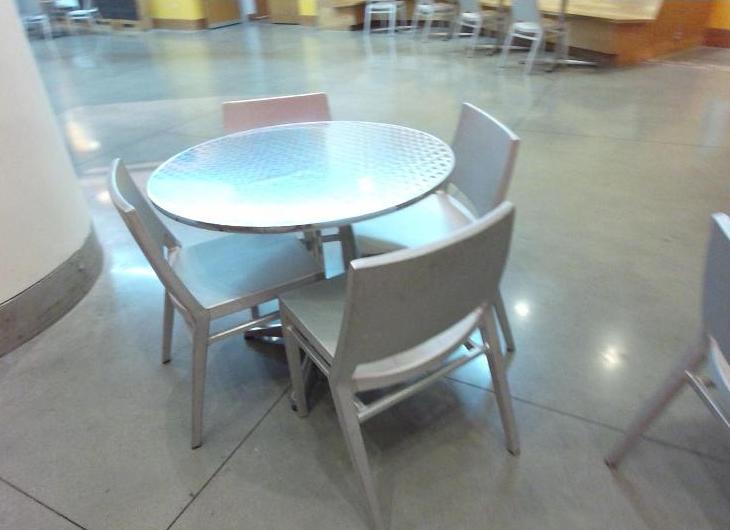}
                \caption{Input}
                % 
                % \caption{\label{fig:anue_image}}        
                
        \end{subfigure}\hfill
        \begin{subfigure}[b]{0.248\textwidth}
                \centering
                \includegraphics[width=\linewidth]{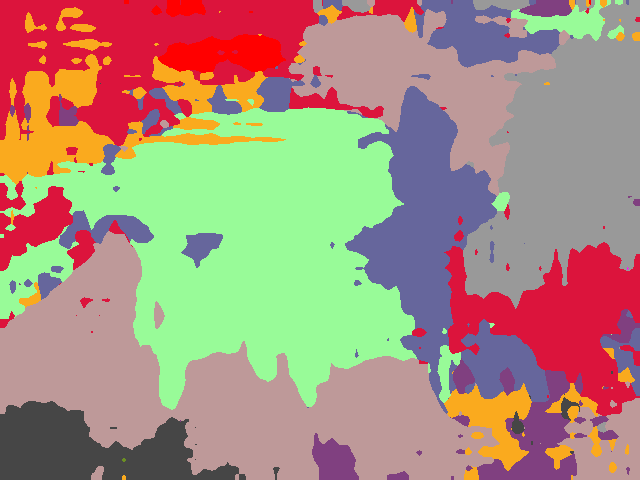}
                \caption{Target-Only}
                % 
                % \caption{\label{fig:anue_bad}}
        \end{subfigure}\hfill
        \begin{subfigure}[b]{0.248\textwidth}
                \centering
                \includegraphics[width=\linewidth]{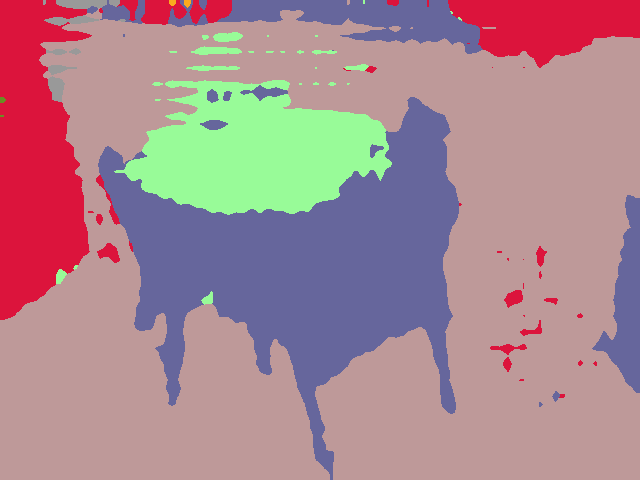}
                \caption{Proposed C2A}
                % 
                % \caption{\label{fig:anue_good}}
        \end{subfigure}\hfill
        \begin{subfigure}[b]{0.248\textwidth}
                \centering
                \includegraphics[width=\linewidth]{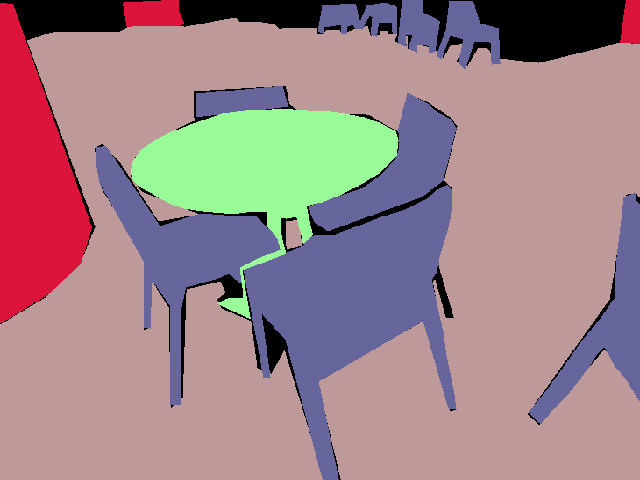}
                \caption{GT Map}
                % 
                % \caption{\label{fig:anue_label}}         
                
        \end{subfigure}\hfill
        \vspace{1em}
        \caption{Qualitative segmentation outputs for examples from the SUNRGB validation set. Compared to a baseline model that is only trained on the few-shot target domain data, the proposed model (C2A) consistently produces better segmentation maps compared to the baselines in all cases.}
        \label{fig:qualitative}
\end{figure*}

\begin{table*}[!t]
  \begin{minipage}{.4\linewidth}{
    \centering
    \includegraphics[width=.9\textwidth]{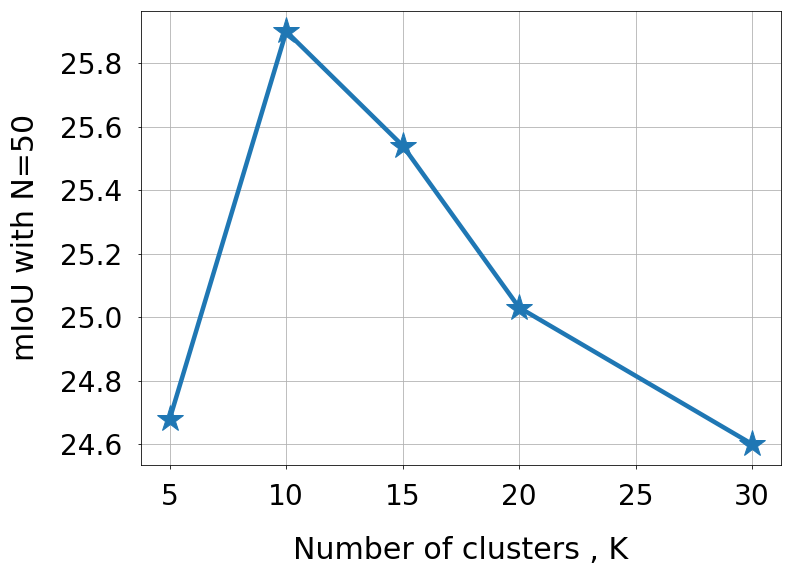}
    \captionsetup{width=\textwidth}
    \caption{
    \textbf{Effect of $K$:}
      Influence of the number of clusters $K$ on the performance of segmentation in the case of $N$=50. 
    \label{fig:clustering_exp}
    }
  }
  \end{minipage}
   ~~~~~~~~~~~~~~~~~~~~~~~~~~~~~~~~~~~
  \begin{minipage}{.35\linewidth}{
    \centering
    \resizebox{\textwidth}{!}{
    % \Huge
    \begin{tabular}{c  c  c}
    \toprule
    Method & SUN  & NYUv2 \\
    \midrule
    Train on SceneNet~\cite{mccormac2017scenenet} & 14.09 & 15.05  \\
    Joint train on SN+GTA & 16.99 & 18.39  \\ 
    \midrule 
    Ours (C2A); K=5 & 21.64 & 22.06 \\
    Ours (C2A); K=10 & 21.89 & 20.27 \\
    Ours (C2A); K=20 & \textbf{22.79} & \textbf{23.08} \\
    % \multicolumn{3}{c}{}\\
    \bottomrule
    \end{tabular}
    }
    \captionsetup{width=\textwidth , type=table}
    \caption{ \textbf{Zero Shot Unsupervised Adaptation:} {Our approach significantly outperforms all the baselines, even in the extreme case of having 0 real target images during training.}
    %   of joint training and finetuning without having any explicit attribute relations. 
    }
    \label{tab:zero_shot}
  }
    
  \end{minipage}
\end{table*}
 \vspace{-8pt}

% \begin{figure}[t!]
%   \begin{minipage}{.45\linewidth}{
%     \begin{center}
%     \includegraphics[width=.9\textwidth]{eccv2020kit/scripts/figures/cluster_fig.png}
%     \end{center}
%     \captionsetup{width=\textwidth}
%     \caption{
%     \textbf{Effect of $K$:}
%       Influence of the number of clusters $K$ on the performance of segmentation in the case of $N$=50. 
%     }
%     \label{fig:clustering_exp}
%   }
%   \end{minipage}
% ~~~~~~~
%   \begin{minipage}{.45\linewidth}{
%     \begin{center}
%     \includegraphics[width=.9\textwidth]{eccv2020kit/scripts/figures/cluster_consistent.png}
%     \end{center}
%     \captionsetup{width=\textwidth}
%     \caption{
%     % \textbf{}
%       The overall performance trends are robust to the choice of the few shot samples selected in the training set. The horizontal axis denotes the different non-overlapping splits used in training.
%     }
%     \label{fig:robust_exp}
%   }
%   \end{minipage}
% \end{figure}

\vspace{-.5em}
\noindent \paragraph{\emph{Role of intermediate bridge domain}}
The intuition behind using an intermediate domain is to ease the adaptation process between the synthetic source domain data and real target domain data, which differ in both the appearance and the label spaces (categories). 
The use of an unlabeled domain bridge leads to no degradation with respect to a direct GTA to SUN adaptation for N = 200 (33.5\% without and 33.4\% with), while leading to a noticeable benefit for N = 50 (25.0\% without and 26.0\% with bridge domain). 

\vspace{-.5em}
\noindent \paragraph{\emph{Comparison with semi-supervised learning methods}} We also compare our work with existing semi-supervised segmentation algorithms in literature, namely AdvSemiSeg~\cite{hung2018adversarial} and universal semi-supervised segmentation~\cite{kalluri2019universal} and include results in \tabref{tab:main_results}. For \cite{hung2018adversarial}, we use the unlabeled and labeled image sets from SUN-RGB to run the experiments. For \cite{kalluri2019universal}, we follow the setting of their paper and use N=(50,200,500,1500) labeled examples from source and target, and use rest of images without annotations. From table~\tabref{tab:main_results}, we note that our method delivers much better performance compared to semi-supervised learning methods for lower amounts of target supervision, as the latter do not leverage rich supervision available from a source domain.

\vspace{-.5em}
\noindent \paragraph{\emph{\textbf{Category-wise Performance}}} We show the classwise mIoU results of our method in \tabref{tab:class_wise_iou} for both cases of using GTA and Synthia as the source dataset. The proposed C2A approach outperforms the baselines, that do not make use of the alignment strategy, on most of the classes. The gains are especially significant on classes like floor, wall and ceiling, which share many geometric as well as semantic properties with categories in Cityscapes and GTA (\figref{fig:qualitative}). For example, the patches corresponding to road in GTA dataset can help to successfully identify the parts of indoor images that correspond to floor since both occur mostly in the lower parts of images and share many other appearance and geometric relations. 
% Similarly, in classes where there is no correspondence to any source category, the performance is retained showing absence of any negative alignment. This supports our initial hypothesis that knowledge transfer across different yet related classes would positively affect the overall performance in cases when target labeled data is very low.

\vspace{-1em}
\noindent \paragraph{\emph{\textbf{$L_c$ and $L_{KL}$}}} For the ablation into clustering losses, we found that removing KL Loss (using only $L_c$) drops performance to 24.24\%, removing clustering loss (using only $L_{KL}$) drops to 23.32\%, while using both these losses gives 25.98\% for the case of N = 50 in \tabref{tab:main_results}.
We can conclude that both the clustering loss as well as the KL divergence loss are necessary as they offer complementary benefits (discussed in Sec 3.3).

% \subsection{Number of clusters}
% \label{subsec:experiment_on_clusters}

\vspace{-1em}
\noindent \paragraph{\emph{\textbf{Number of clusters}}} An important aspect of our formulation is the choice of number of clusters $K$ for the clustering approach in~\eqref{eq:clustering_loss}. From \figref{fig:clustering_exp}, K=10 clusters works well for the case of $N=50$. This concurs well with our intuition that a small value of $K$, like 5, would adversely affect the discriminative performance of the original task while a large value of $K$, like 20, would not encourage the semantic transfer we are after, as even related categories might form distinct clusters with no overlap 

% \subsection{Unsupervised Zero Shot Adaptation}

\vspace{-1em}
\paragraph{\emph{\textbf{Unsupervised Zero Shot Adaptation}}}

With a slight modification, our approach also works well for the problem of zero shot adaptation, when no (labeled, or unlabeled) examples are available from the target domain. 
% This setting is similar to the previously proposed RT-IRT based adaptation~\cite{wang2019conditional}, but in a more generalized framework with completely different tasks extended to semantic segmentation. 
For learning some task specific information, we assume availability of synthetically generated samples from the target domain. In our case, we render artificial indoor scenes from SceneNet~\cite{mccormac2017scenenet} dataset and use it in conjunction with our approach. We use this synthetically generated data from SceneNet instead of SUNRGB in the formulation, in \eqref{eq:supervised_loss} and \eqref{eq:clustering_loss} instead of $\D_t$. 

We compare our approach against the baselines where we use a classifier trained on SceneNet directly on SUNRGB. We report the results in \tabref{tab:zero_shot}, and our method which jointly optimizes a clustering objective along with an adversarial objective performs much better than the baselines that use only synthetic images from SceneNet. We believe that is due to the complementary knowledge that the network is able to infer through our approach.
% by learning the discriminative features from the SceneNet dataset, and more abstract lower level representations from GTA, resulting in superior performance without even using any real images from the target domain. 
Our method even improves upon plain joint training on labeled GTA and SceneNet datasets from $\sim17\%$ to $\sim22.8\%$, indicating that the benefit is not only due to increase in labeled data, but due to alignment as well.
% To verify the the fact that the benefit is not only due to the increase in amounts of labeled data, we also compare our approach against joint training the model on labeled GTA and SceneNet datasets, and report considerable absolute improvement  using $K=20$ clusters. 
Additionally, a network which learns without any real data from target domain should ideally generalize well to any similar datasets. Indeed, we observe from \tabref{tab:zero_shot} that the improvement on performance is not restricted to SUNRGB alone, but also observed on NYUv2~\cite{silberman2012indoor} validation set across all the baselines.

\section{Conclusion}

We introduce C2A, a clustering based approach called C2A to study the most general, yet largely understudied setting of adaptation between domains with non-overlapping label spaces for feature alignment across source and target datasets with disjoint labels. C2A encourages positive alignment of visually similar feature representations while preventing negative transfer. We experimentally verify the effectiveness of our approach on the task of outdoor to indoor adaptation for semantic segmentation and demonstrate significant improvements over existing approaches and prevalent baselines in both fewshot and zeroshot adaptation settings. 

\noindent {\bf Acknowledgements} We thank NSF CAREER 1751365, NSF Chase-CI 1730158, Google Award for Inclusion Research and IPE PhD Fellowship.

% \clearpage

\bibliographystyle{ieee_fullname}
\bibliography{main}

\begin{appendix}

\section{Cluster Initialization}

The cluster centers are initialized using networks pretrained on the limited labeled data. Specifically, we use the same architecture as described in the paper to train a model on the labeled source data $\D_s$ as well as sparsely labeled target data $\D_t^l$ using pixel level cross entropy loss. We then pass the unlabeled images from the target $\D_t^u$ and collect all the encoder maps corresponding to all the images. Each encoder map is of size $(H/8 , W/8 , 2048)$ for our ResNet-101 backbone. To match the dimension of the FTN output, which is $128$ in our case, we apply PCA over these feature vectors to reduce their dimension. Then, a clustering is performed using the classical k-means objective with $K$ cluster centers, and the resulting centers are used to initialize $\mu_k's$ in the downstream adaptation approach. 

\paragraph{ \emph{\textbf{Efficient Computation of Centers}}} In traditional k-means, the centers $\mu_k$ are calculated using an iterative algorithm consisting of cluster assignment and centroid computation repeated until convergence. We mention a couple of issues persistent with this approach. Firstly, for dense prediction tasks like semantic segmentation, the encoder map consists multiple feature vectors which correspond to different patches of the input image. For example, an encoder map of size $(H',W')$ has $H' W'$ vectors of size $f_e$. Performing k-means over these vectors collected over all images over all the tasks would demand huge storage and computation requirements. Secondly, switching between gradient based training of network parameters and iterative computation of cluster centers after every few iterations would lead to an inefficient procedure that is not end-to-end trainable. To counter these limitations, we follow the idea proposed in ~\cite{han2019learning} and include $\mu_k$ as trainable parameters in the network, and update them after each iteration based on the gradients received from $\Loss_C$. 

\subsection{Datasets} 

For the source dataset $\D_s$, we use synthetic images from the driving dataset GTA~\cite{richter2016playing}. GTA consists of 24966 images synthetically generated from a video game consisting of outdoor scenes with rich variety of variations in lighting and traffic scenes. We also show results using the SYNTHIA-RAND-CITYSCPAES split from Synthia~\cite{ros2016synthia} dataset, which consists of 9600 synthetic images with labels compatible with Cityscapes. 
For the target dataset $\D_t$, we use real images from SUN-RGBD~\cite{song2015sun} consisting of images from indoor scenes. SUN-RGBD consists of 5285 training images and 5050 validation images containing pixel level labels of objects which frequently occur in an indoor setting like chair, table, floor, windows etc. We use the 13 class version from~\cite{mccormac2017scenenet}. The background class is ignored during training and evaluation.
% We believe that this choice of datasets is significant because knowledge transfer between indoor and outdoor images for semantic segmentation was never explored in literature before. 
Additionally, we use the 2975 training images from Cityscapes~\cite{cordts2016cityscapes} dataset, which consists of outdoor traffic scenes captured from various cities in Europe, as the unlabeled auxiliary domain $\D_a$. Cityscapes shares its semantic categories with GTA, so that the variation between $\D_s$ and $\D_a$ is only due to synthetic and real appearance, while $\D_s$ and $\D_t$ have many low-level as well as high-level differences. 
% The semantic categories are shared between Cityscapes and GTA.

% Furthermore, we also test out approach in a complementary, yet related, setting of zero shot adaptation. Specifically, we assume that we do not have any labeled or unlabeled images from the target dataset. Instead, we assume we have access to synthetically generated images that share categories with the target dataset. To validate this approach, we use all the labeled images from SceneNet~\cite{mccormac2017scenenet} as the target dataset, and show the generalization capabilities to datasets not seen before.
% We report all our results on the validation set of SUN-RGBD datasets. 

\section{Training Details}
We use the DeepLab~\cite{chen2017deeplab} architecture with a resnet-101 backbone for the encoder framework $\E$. For the task-specific decoder $\Dec$, we use an ASPP convolution layer followed by an upsampling layer. The architecture of discriminator $\Disc$ is similar to DC-GAN~\cite{radford2015unsupervised} with four $4 \times 4$ convolution layers, each with stride 2 followed by a leaky ReLU non-linearity. The feature transformation module $\FTN$ is a $1 \times 1$ convolution layer with output channels equal to the embedding dimension, which is fixed as $f_e=128$ for all the experiments. We use a default value for $\lambda_{adv}=0.001$. Following~\cite{xie2018learning}, to suppress the noisy alignment during the initial iterations, we set $\lambda_c = \tfrac{2}{1+e^{-10*\delta}}-1$, where $\delta$ changes from 0 to 1 over the course of training. The backbone architecture is trained using SGD objective, with an initial learning rate of $2.5 \times 10^{-4}$. For training the cluster centers, we follow a similar learning rate decay schedule, but start with a smaller learning rate of $2.5 \times 10^{-5}$. This is because the cluster centers are already initialized using networks trained on the labeled data, and we would ideally like the centers to not drift too far away from their initial values.

% Code and models will be publicly released.
% More training details in the supplementary. 

\paragraph{\emph{\textbf{Metric}}} We use the mean intersection over union (mIoU), as the performance comparison metric. IoU per class per image is defined by
\begin{equation}
    mIoU = \frac{TP}{TP+FP+FN}
\end{equation}
\noindent where TP,FP,FN are the true positive, false positive and false negative predictions in an image respectively. mIoU is the average IoU of all classes across all images in the validation set. We use the 5050 validation images in the SUN-RGB dataset to report our results.

\paragraph{Baselines} For the baselines we compared against, which are \cite{tsai2018learning} and \cite{luo2017label}, we extend those approaches to suit our task of semantic segmentation across disjoint labels. That is, we use feature space adaptation in \cite{tsai2018learning} and use prototype base alignment for \cite{luo2017label}. These are indicated by \textit{AdaptSegNet*} and \textit{LET*} in the main paper.

\begin{figure}[t!]
  \begin{minipage}{.45\linewidth}{
    \begin{center}
    \includegraphics[width=.9\textwidth]{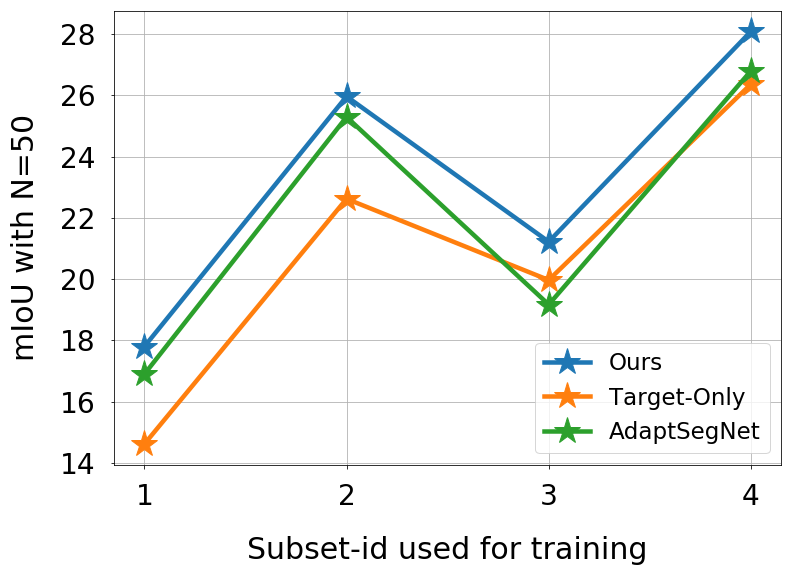}
    \end{center}
    \captionsetup{width=\textwidth}
    \caption{
      The overall performance trends are robust to the choice of the few shot samples selected in the training set. The horizontal axis denotes the different non-overlapping splits used in training.
    }
    \label{fig:robust_exp}
  }
  \end{minipage}
~~~~~~~
  \begin{minipage}{.48\linewidth}{
    \begin{center}
    \includegraphics[width=.9\textwidth]{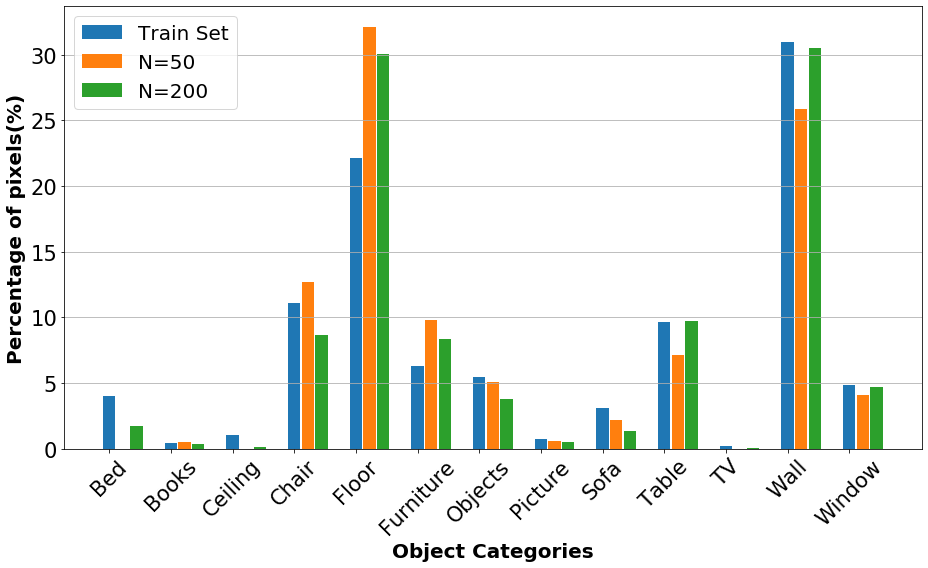}
    \end{center}
    \captionsetup{width=\textwidth}
    \caption{
      The label distribution of the selected few shot data is consistent with the global label distribution.
    }
    \label{fig:label_dist}
  }
  \end{minipage}
\end{figure}
% \section{Training Details}

% We use a Deeplab-V3 base model with ResNet-101 backbone as the encoder network. The training is run for 65000 iterations, and the learning rate is adjusted according to polynomial decay before each iteration starting from an initial learning rate of $2.5 \times 10^{-4}$. Since the decoder and the feature transfer network are trained from scratch as opposed to the encoder which is initialized using weights pretrained on MS-COCO, we use a learning rate of $2.5 \times 10^{-3}$ on the deocoder and the FTN. The encoder, decoder and the FTN are jointly trained using an SGD optimizer with a momentum of $0.9$ and a weight decay factor of $0.0005$. The discriminator is trained with a similar learning rate schedule with an initial learning rate of $10^{-4}$ using an Adam optimizer. The batch normalization parameters, which are initialized from the pretrained network, are not updated during the training phase. 

% \input{eccv2020kit/scripts/figs_consistency}

\section{Label Distribution}

For training the model using the proposed approach, we require $N$ labeled examples from the target SUN-RGB dataset. 
% In all our experiments, we choose the first $N$ examples from the official training set as the labeled target dataset $\D_t^l$. 
Since no official split is available for a few shot setting like ours, we randomly choose a subset of train examples as the few shot examples $\D_t^l$ and fix this set over all the ablation studies. Through \figref{fig:robust_exp}, we demonstrate that our clustering based approach is robust towards the particular subset used, and the method outperforms the baselines as well as the adversarial approach irrespective of the subset used. However, the values vary over a wide range due to the disparity in the particular images used, which underlines the importance of establishing a standard few shot learning benchmark datasets for semantic segmentation as a future work.
We also show the label distribution in our setting compared to the global distribution from the complete train set in \figref{fig:label_dist} for $N=50$ and $N=200$. The percentage of pixels of each class remain the same even in our few shot settings, except for classes like \textit{train} and \textit{sofa}, which are very scarcely present.
Similar to any semantic segmentation task, the label distribution is not uniform across all the classes in the images. We show the label distribution of the chosen samples in \figref{fig:label_dist} for $N=50$ and $N=200$ when compared to the total label distribution.

\end{appendix}
\end{document}